\documentclass[a4paper,10pt,twocolumn]{article}
\usepackage[a4paper,right=1.5cm,left=1.5cm,top=2cm,bottom=2.5cm,headsep=0cm,footskip=0.5cm]{geometry}
\usepackage[english,activeacute]{babel}

\usepackage[utf8]{inputenc}
\usepackage{enumitem}
\usepackage{amsmath}
\usepackage{times}
\usepackage{graphicx}
\usepackage{float}
\usepackage{csquotes} 
\usepackage{multirow}
\usepackage{xcolor}
\usepackage{colortbl}
\usepackage[colorlinks=true,allcolors=blue]{hyperref}
\usepackage{titlesec}
\usepackage{subcaption}
\usepackage{float}
\usepackage[font=small,labelfont=bf,justification=justified,singlelinecheck=false]{caption}
\usepackage{adjustbox}
\usepackage{comment}

\titleformat{\section}{\large\bfseries}{\thesection}{1em}{} 
\titleformat{\subsection}{\small\bfseries}{\thesubsection}{1em}{} 
\titleformat{\subsubsection}{\small\bfseries}{\thesubsubsection}{1em}{} 
\titlespacing{\section}{0pt}{\parskip}{\parskip}
\titlespacing{\subsection}{0pt}{\parskip}{\parskip}

\renewcommand\thesection{\arabic{section}.}
\renewcommand\thesubsection{\arabic{section}.\arabic{subsection}.}

\title{\centering \LARGE \textbf{Enhancing Whole Slide Image Classification through Supervised Contrastive Domain Adaptation}}

\author{\large Ilán Carretero$^1$, Pablo Meseguer$^{1,2}$, Rocío del Amor$^{1}$, Valery Naranjo$^{1,2}$
\\
\\
\small $^1$ HUMAN-tech, Universidad Politécnica de Valencia (UPV), Valencia, España, \{ilcarjuc, pabmees, madeam2, vnaranjo\}@upv.es \\
\small $^2$ Valencian Graduate School and Research Network for Artificial Intelligence (valgrAI), Valencia, España. \\
}
\date{} 

\begin{document}	

\maketitle
\thispagestyle{empty}

\begin{center}
\large
\textbf{Abstract}
\end{center}
\small
\textit{Domain shift in the field of histopathological imaging is a common phenomenon due to the intra- and inter-hospital variability of staining and digitization protocols. The implementation of robust models, capable of creating generalized domains, represents a need to be solved. In this work, a new domain adaptation method to deal with the variability between histopathological images from multiple centers is presented. In particular, our method adds a training constraint to the supervised contrastive learning approach to achieve domain adaptation and improve inter-class separability. Experiments performed on domain adaptation and classification of whole-slide images of six skin cancer subtypes from two centers demonstrate the method's usefulness. The results reflect superior performance compared to not using domain adaptation after feature extraction or staining normalization.} 

\normalsize

\section{Introduction}

Deep Learning (DL) models developed in the medical field must be robust to changes in the distributions of the input data promoting their generalization. The domain shift is defined as the difference between source and target domain data due to variations in image characteristics \cite{stacke2020measuring}. In digital pathology, domain shift is manifested as variability in terms of staining and digitalization protocols, specifically between different hospitals. Training robust DL models that tackle domain shift and generalize well across multiple domains remains crucial to provide accurate patient diagnosis.

Staining normalization has been the most investigated approach to pursue domain generalization in histopathological images. To this end, methods based on H\&E component separation \cite{macenko2009method} or image-to-image translation with generative adversarial networks \cite{zhu2017unpaired} have been proposed to create a unified color space. However, these methods usually rely on target images and are not transferable to problems assessing tumors located in different tissue types. Other approaches have employed unsupervised contrastive techniques for histopathological image domain adaptation \cite{vray2024distill}. Nevertheless, these methods are highly demanding in terms of computational effort and volume of images. These limitations are particularly critical in the field of histopathological imaging since a large sample number to train DL-based models is not always available.

Based on these observations, in this work, we design an efficient domain adaptation method to handle the variability related to the staining and scanning of the whole-slide images across multiple centers. The proposed framework, termed Supervised Contrastive Domain Adaptation (SCDA), addresses the representational shift of foundation models for the slide-level prediction of multiple skin cancer subtypes in a multi-center dataset. Our domain adaptation technique is based on supervised contrastive learning \cite{khosla2020supervised} with an embedded constraint during model training forcing the addition of samples from different centers. The low computational burden of the network coupled with the extensibility to the few-shot learning paradigm promotes efficient adaptation to solve the task at hand. 

The main contributions are summarized as follows: Development of a methodology based on supervised contrastive learning for cross-center domain generalization, extension of our SCDA method to the few-shot learning paradigm, obtaining efficient domain adaptations with fewer exemplars of a hospital not previously covered in the training phase and evaluation of the quantitative and qualitative classification results compared to non-normalization and stain normalization, demonstrating significant improvement.

\section{Methodology}
\begin{figure}[h]
\centering
        \includegraphics[width=0.45\textwidth]{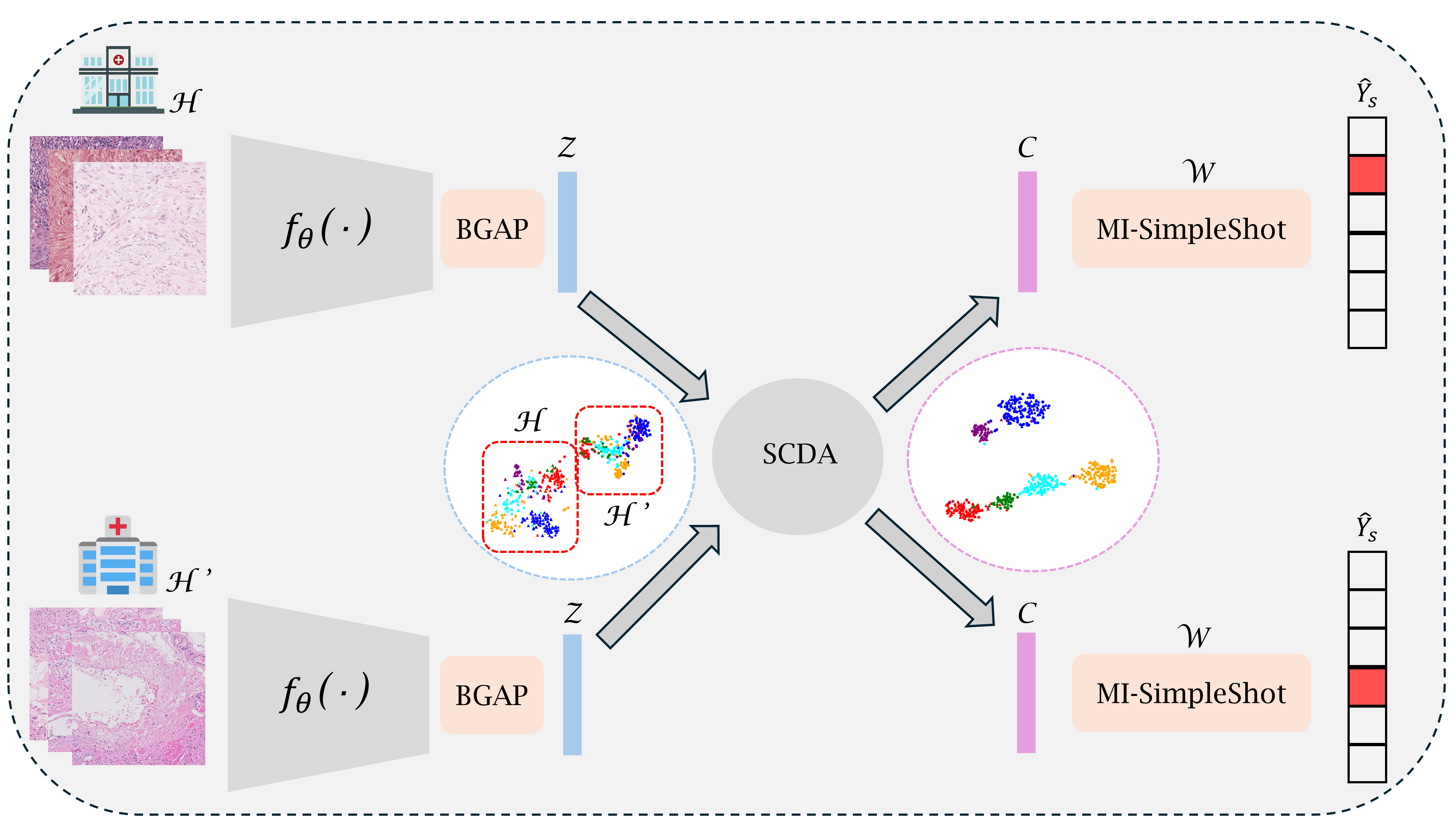}
    \caption{\textit{Method overview. In this article, we address inter-center domain shifts through supervised contrastive learning. Concretely, a constraint is introduced in the model training by forcing samples of the same class to be closely spaced. The inclusion of this condition strengthens inter-class clustering and removes inter-center variability.}}
    \label{fig:SCDA}
    \vspace*{-5mm}
\end{figure}

An overview of the proposed method is illustrated in Fig. \ref{fig:SCDA}. The problem formulation and the different components implemented are described in the following. 

\subsection{Problem formulation}
In the multiple instance learning (MIL) paradigm, an arbitrary number $N$ of instances $x$ are grouped into bags $X = \{x_n\}^N_{n=1}$. Each bag is assigned to a class $S$ that is mutually exclusive from all other classes. Thus, in a multi-class environment, the assignment of $X$ to a class is denoted as $Y_S \in \{0,1\}$. In the embedding-MIL approach, the goal is to obtain the WSI-level prediction using a bag-level representation $Z$ of the features extracted at the patch level. Let $H_1$ and $H_2$ be different centers with different datasets but the same $S$ classes. The objective is to find a transformation function $\mathcal{T}$ that drives $Z_{H_1}$ and $Z_{H_2}$ to a common feature space $C$, such that $T:Z_{H_1} \cup Z_{H_2} \rightarrow C$. Consequently, by finding $\mathcal{T}$, we can define a prediction function $\mathcal{F}:C \rightarrow Y_{H'}$ for the bag-level multi-class classification problem of $H'$, using only the transformed representations $\mathcal{T}(Z_{H})$ and labels $Y_H$.

\subsection{Deep slide-level representation}
MIL approaches heavily rely on feature extraction at the patch level to subsequently obtain a global representation of the bag. Let us denote a feature extractor $f_{\theta}(\cdot):\mathcal{X}\rightarrow\mathcal{F}$, which projects the instances $x \in \mathcal{X}$ to a lower-dimensional manifold $z \in \mathcal{F} \subset R^{d}$, where $d$ is the embedding dimension. Then, we utilize a non-trainable aggregation to obtain the slide-level representation $Z$ by performing batch global average pooling (BGAP) across all the instances within a slide. Note that both the patch-level feature extraction and the WSI embedding aggregation do not require any parameter update, promoting the efficiency of our proposed method for domain adaptation under a few-shot learning scenario.

\subsection{Supervised Contrastive Domain Adaptation}
Contrastive learning is a technique that learns representations by contrasting similar and dissimilar data pairs to improve model discrimination. In this paradigm, supervised contrastive learning is established as a method capable of taking advantage of label information. Mathematically, the supervised contrastive loss for univiewed batch (typically, in contrastive learning one has multiviewed batches where there is more than one view for a single sample) can be defined as:

\begin{align}
   \mathcal{L} = \sum_{i \in I} \mathcal{L}_i = \sum_{i \in I} -\log \frac{\exp(\pmb{z_i} \cdot \pmb{z_p} / \tau)}{\sum_{a \in A(i)} \exp(\pmb{z_i} \cdot \pmb{z_a} / \tau)}
 \label{eq:scl}
\end{align}

where $\pmb{z_i}$ is a representation of sample $i$, $\pmb{z_p}$ is a representation of a sample $j\neq i$ that belongs to the same class, $A(i)$ is the set of all representations of the batch and $\tau$ is a temperature parameter that scales the similarity between the representations. A cross-domain constraint is added to successfully apply supervised contrastive learning to the domain adaptation, such that:
\vspace*{1mm}
\footnotesize
\begin{align*}
    \forall S, \forall B, \exists i \in I_H \cap B, \exists j \in I_{H'} \cap B \; \text{s.t.} \; \text{class}(i) = \text{class}(j) = S
\end{align*}
\normalsize
being $I$ the set of instances of a center $H$ and $B$ the batch where this restriction applies. Equation \eqref{eq:scl} encourages sample representations of the same class to cluster together and be further distanced from representations of other classes. Consequently, include the cross-domain constraint forces to consider inter-center variations for the same class. It is worth mentioning that applying \eqref{eq:scl} with the cross-domain constraint yields new slide-level $C$ representations, which are the transformations of the $Z$ representations.

\subsection{MI-SimpleShot}
The classification step in MIL frameworks usually constructs a multilayer perceptron (MLP) and a softmax activation function to obtain bag-level scores from the WSI embedding. However, the MLP optimization through minimization of the cross-entropy loss does not handle class imbalance and could lead to overfitting, hampering model generalization. For that purpose, our framework adapts the prompt-based classification approach named MI-SimpleShot \cite{chen2023general}. We utilize the transformed bag-level embeddings $C$ to construct the class prototypes ($W$) and the prediction of each slide is finally assessed by leveraging the largest similarity to the prototypes of each class, such as:
\begin{equation}
\label{eq_classification}
\hat{Y} = \text{argmax} (W^T C)
\end{equation}

\section{Experimental setting}
\subsection{Dataset}

We resort to a multi-center dataset to evaluate the SCDA method for cross-center domain adaptation for slide-level classification \cite{del2023annotation}. The dataset contains up to 608 skin WSI from two different centers: Hospital Clínico of Valencia (HCUV) and the Hospital Universitario San Cecilio (HUSC) of Granada. See Table \ref{tbl_dataset} for a summary of the number of slides available for each neoplasm within each center.

\begin{table}[ht]
\setlength\tabcolsep{2 pt}
\small
\begin{center}
\begin{tabular}{lccc}
\hline
\multicolumn{1}{l}{}{} & \textbf{HCUV}  & \textbf{HUSC} & \textbf{HCUV $\cup$ HUSC}\\
\hline
\textbf{Leiomyoma (lm)}    & $ 31 $    & $ 71 $ & $ 102 $\\
\textbf{Leiomyosarcoma (lms)}   &  $ 25 $    & $ 26 $  & $ 51 $\\
\textbf{Dermatofibroma (df)}   &  $ 73  $     & $ 95 $ & $ 168 $\\
\textbf{Dermatofibrosarcoma (dfs)}    & $ 17 $    & $ 44 $  & $ 61 $\\
\textbf{Spindle-cell melanoma (mel)}    & $ 49 $    & $ 73 $  & $ 122 $\\
\textbf{Atypical fibroxanthoma (fxa)}   & $ 44 $    & $ 60 $  & $ 104 $\\
\hline
\textbf{All classes}   & $ 239 $    & $ 369 $  & $ 608 $\\
\hline
\end{tabular}
\caption{\textit{Description of the dataset used containing WSI of skin cancer subtypes from two centers: Valencia (HCUV) and Granada (HUSC).}}
\vspace*{-5mm}
\label{tbl_dataset}
\end{center}
\end{table}

As stated before, this work pretends to address the generalization ability of slide-level models through multiple centers that could present domain shifts regarding staining and scanning. In Section \ref{sec:results}, we provide more insights into the domain shift at the representation level.

\subsection{Implementation details}

To accommodate the MIL paradigm, the WSIs were cropped into 512 squared pixel patches at 10x magnification with a 50\% overlap.  We utilized the Pathology Language Image Pretraining (PLIP) model for feature extraction at the patch level \cite{huang2023visual}. PLIP is a foundation model trained with vision language supervision using a dataset of histological images with their paired textual descriptions retrieved from Twitter. Foundation models trained with large-scale in-domain datasets promise to obtain features with strong representation quality that can be used to perform multiple downstream tasks. As stated in Section \ref{sec:results} this work is partially motivated by the findings about representation shift observed when using the PLIP image encoder for feature extraction (see Fig. \ref{fig:tsne_og}).

Our framework pretends to assess the domain shift to improve cross-hospital generalization in few-shot scenarios, thus promoting model efficiency and handling limitations to high computational costs related to WSI. We investigated the few-shot adaptation under different numbers of training samples ($k = \{2,4,6,8,10\}$) per class for domain adaptation. Experimentation is run through five different seeds to measure the variability inherent to random few-shot selection. All models are evaluated regarding balanced accuracy (BACC) to consider unbalanced classes equally. The dataset was partitioned with an 80\% rate for training and 20\% for testing. Note that these sets remain fixed through all the experiments for a fair comparison of the different configurations of cross-hospital prediction. 

\section{Results}
\label{sec:results}

\subsection{Quantitative evaluation}

The classification results obtained are shown in Table \ref{tab:results}. The balanced accuracies obtained by applying MI-SimpleShot on the PLIP and BGAP clustered representations are reasonably successful when predicting samples from the same dataset as the trained dataset or training both datasets ($0.69$ for HCUV, $0.80$ for HUSC and $0.76$ for both centers). However, a considerable drop in performance is observed when we infer the untrained hospital ($0.54$ for HCUV and $0.51$ for HUSC). As seen in the qualitative results (see Fig. \ref{fig:tsne_og}), the inter-hospital differentiation in the PLIP representations is very prominent. This means that the generalization capacity of the model in the face of domain shifts is partially limited. To solve the inter-hospital differentiation problem, Macenko \cite{macenko2009method} stain normalization was applied. Nevertheless, since the color of each WSI has a direct dependence on the cellular response to staining, eliminating this information results in a decrease in the classification performance compared to not applying Macenko normalization for domain generalization. When applying our approach, a considerable increase in all test metrics is observed with respect to the two previous approaches. This is due to the fact that our SCDA method removes inter-hospital variability and clusters the skin cancer subtypes in a more condensed way through supervised contrastive loss (see Fig. \ref{fig:tsne_scl}).

\begin{table}[ht]
\centering
\setlength\tabcolsep{1.5 pt}
\begin{adjustbox}{width=0.47\textwidth}
\begin{tabular}{ccccc}
\hline
\textbf{Method}        & \textbf{Training centers} & \multicolumn{3}{c}{\textbf{Testing centers}} \\ \hline
                       &                           & HCUV    & HUSC    & HCUV $\cup$ HUSC   \\ \hline 
                       & HCUV                      & 0.69    & 0.51    & 0.58          \\
                       & HUSC                      & 0.54    & 0.80    & 0.70          \\
\multirow{-3}{*}{PLIP \cite{huang2023visual}} & HCUV $\cup$ HUSC               & 0.71    & 0.80    & 0.76          \\ \hline 
Macenko \cite{macenko2009method} + PLIP \cite{huang2023visual}       & HCUV  $\cup$ HUSC              & 0.60    & 0.73    & 0.68          \\ \hline 
\rowcolor[HTML]{EFEFEF} 
PLIP \cite{huang2023visual} + SCDA            & HCUV $\cup$ HUSC               & \textbf{0.88}    & \textbf{0.93}    & \textbf{0.91}          \\ \hline
\end{tabular}
\end{adjustbox}
\caption{\textit{Quantitative comparison of the results obtained in the different test sets. The metric presented is the balanced accuracy (BACC). HCUV $\cup$ HUSC means that HCUV and HUSC data sets have been merged for training or inference.}}
\vspace*{-5mm}
\label{tab:results}
\end{table}

\begin{figure}[ht]
     \centering
     \begin{subfigure}[b]{0.22\textwidth}
         \centering
         \includegraphics[width=\textwidth]{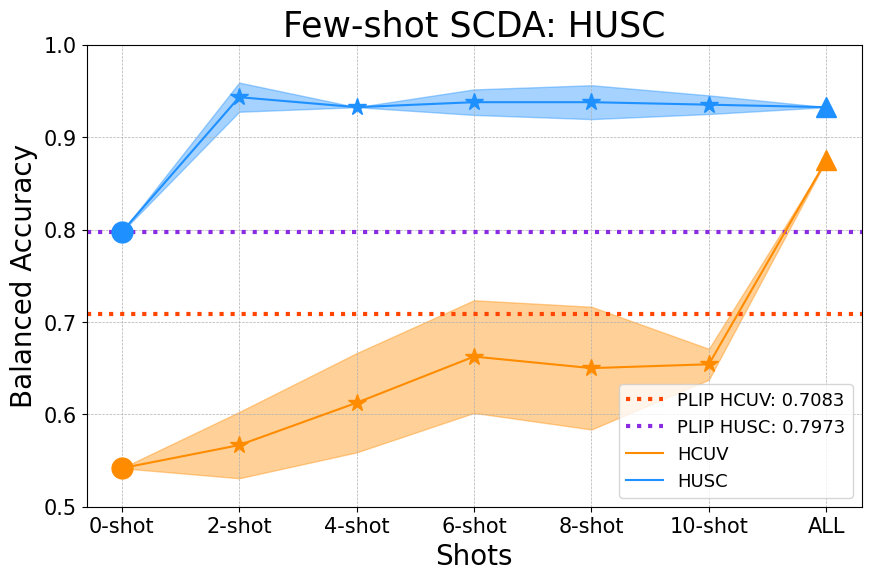}
         \caption{\textit{Few-shot regime. Training set: HUSC}}
         \label{fig:y equals x}
     \end{subfigure}
     \hfill
     \begin{subfigure}[b]{0.22\textwidth}
         \centering
         \includegraphics[width=\textwidth]{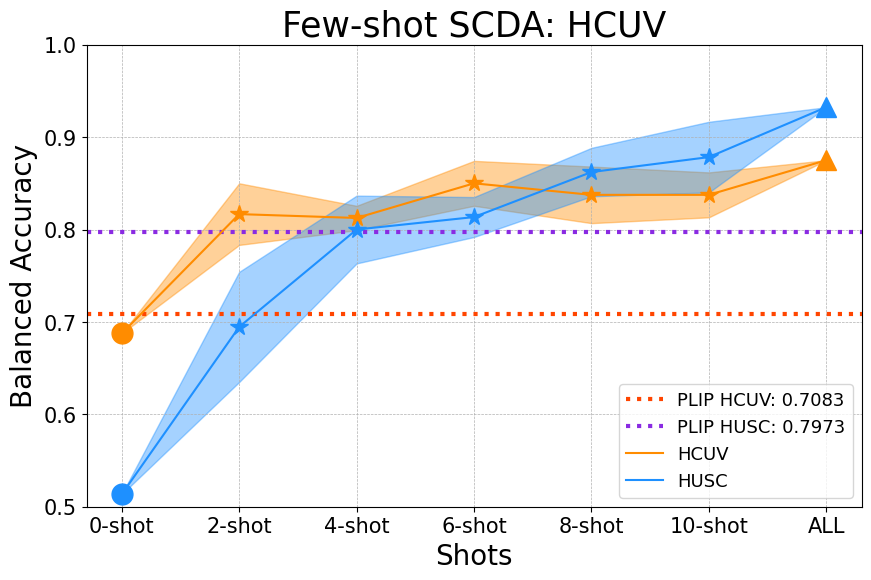}
         \caption{\textit{Few-shot regime. Training set: HCUV}}
         \label{fig:five over x}
     \end{subfigure}
        \caption{\textit{In-depth study of SCDA under the few-shot learning paradigm. In (a) HUSC train set is used, and HCUV samples are added. In (b), the reverse procedure is evaluated. The results of the HUSC test set are shown in blue and those of HCUV in orange. The starting point on the curves refers to the 0-shot case, where only the PLIP and BGAP feature extractors are used to predict. The final triangle in the curves refers to the case where both databases are trained. The orange horizontal dotted line indicates the results for the HUCV test set trained with the two databases without using SCDA. The purple dotted line refers to the same but with the HUSC dataset.}}
        \label{fig:few_shot_SCL}
        \vspace*{-0mm}
\end{figure}

In light of the results obtained in the different test sets of each hospital using SCDA, we propose to evaluate our method in a few-shot learning scenario. The results obtained for different numbers of training samples are illustrated in Fig. \ref{fig:few_shot_SCL}. When training with the HUSC data, it is seen that for the HUSC test set the performance is maintained regardless of the new knowledge introduced with the HUCV samples. With respect to inference on the HUCV test set, it is found that there is a significant improvement in inference performance as the number of shots is increased. With an increase of $\sim10\%$ in the balanced accuracy from 0-shot to 10-shot, the usefulness of SCDA for cross-hospital slide-level classification based on just a few reference images is evident. 

If we analyze the few-shot learning scenario by training with the HCUV dataset and adding $k$-samples from the HUSC training set, a significant improvement in the metrics can be seen. Thus, in the HUCV test set, the performance remains consistent regardless of the number of shots. In contrast, in the HUSC test set, the performance as the number of shots increases considerably improves. In this way, from 4-shots onwards, it can be seen how the application of SCDA improves the metrics obtained in the HUSC test set compared to using the whole set of train without using our SCDA method. Regarding the increase from 2-shot to 10-shot, it can be seen that it is $\sim15\%$. It should also be mentioned that from 8-shot onwards, comparable metrics are achieved when including the whole HUSC training set (ALL). This proves that the few-shot learning approach with SCDA is really powerful since the generalization capacity obtained with a few samples from a different domain is highly significant.

\subsection{Qualitative evaluation}

\begin{figure}[ht]
     \centering
     \begin{subfigure}[b]{0.22\textwidth}
         \centering
         \includegraphics[width=\textwidth]{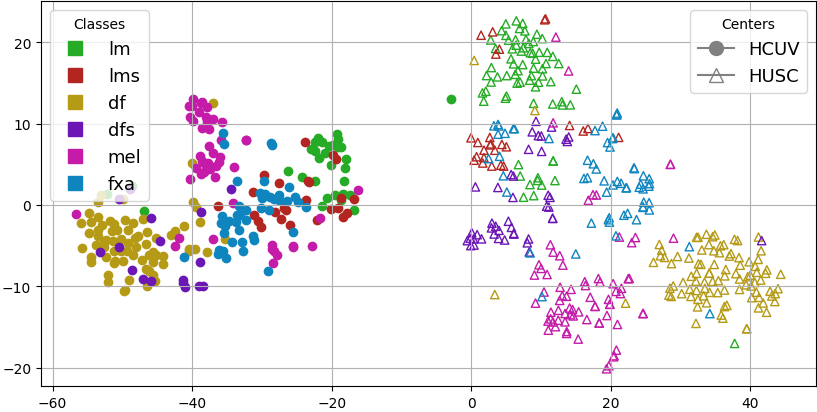}
         \caption{\textit{Feature embeddings 2D representation}}
         \label{fig:tsne_og}
     \end{subfigure}
     \hfill
     \begin{subfigure}[b]{0.22\textwidth}
         \centering
         \includegraphics[width=\textwidth]{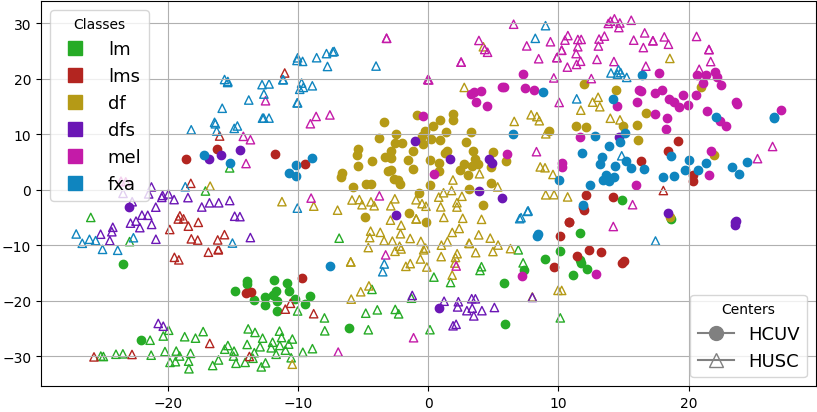}
         \caption{\textit{Feature embeddings with Macenko normalization}}
         \label{fig:tsne_ma}
     \end{subfigure}
     \hfill
     \begin{subfigure}[b]{0.22\textwidth}
         \centering
         \includegraphics[width=\textwidth]{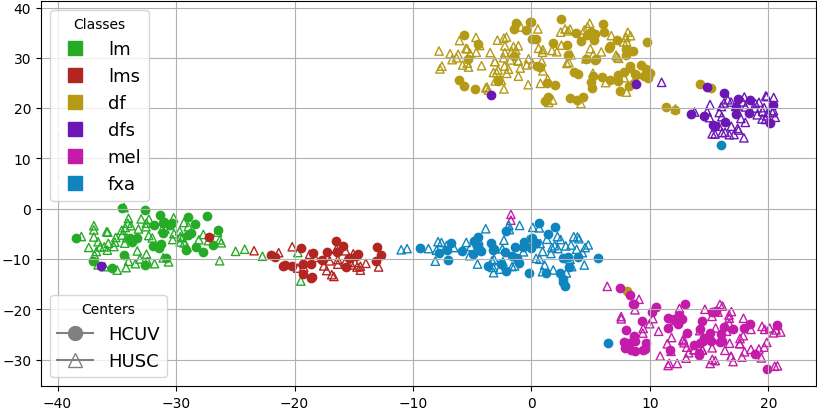}
         \caption{\textit{Feature embeddings with SCDA}}
         \label{fig:tsne_scl}
     \end{subfigure}
        \caption{\textit{2D t-SNE representations of feature embeddings extracted by PLIP and aggregation via BGAP. The subfigures illustrate (a) Representation for the embeddings without any added processing. (b) Representation of the embeddings obtained with PLIP and BGAP by previously applying the Macenko normalization method. (c) Representation of the embeddings by applying SCDA.}}
        \label{fig:tsnes}
        \vspace*{-5mm}
\end{figure}

To appreciate the representation shift between the domain of the two centers, we represent in Fig. \ref{fig:tsnes} the 2D t-SNE representations for the original PLIP embeddings and the two approaches for cross-hospital domain adaptation. The dimensionality reduction of the BGAP embeddings of the PLIP shows a large domain shift between centers. However, it is noteworthy the strong representation learning of the feature extractor as the cluster of each class within a center are quite separated from the others. This visualization helps to understand the poor generalization ability (see Table \ref{tab:results}) of the initial model as the latent spaces separated due to bias between centers. 

Afterward, we compare the WSI-level embeddings after applying stain normalization to a target image of the HUSC dataset. Although the stain bias is mitigated for certain classes (lm, df), others (fxa, dfs) remain separated in the deep latent distribution. Moreover, the compactness of the samples within each class has been downgraded, suggesting that the color can provide valuable information to differentiate between tumors with different tissues. Finally, our SCDA approach across different centers not only mitigates the stain bias of the foundation model, but also enhances the similarity between samples of each neoplasm.

\section{Conclusion}

The intrinsic staining and digitization variability in the field of histopathological imaging, significantly affects the implementation of robust DL models. Solving domain shifts without loss of performance is essential for DL model generalization and implementation. In this work, we introduce a new domain adaptation method applied to histopathological images of a multicenter dataset. By using the supervised contrastive learning technique together with a cross-domain constraint, it is possible to transform features extracted from a foundational model into a generalized domain. Experiments performed using the HCUV and HUSC dataset demonstrate the power of the implemented method. The metrics obtained considerably outperform cross-domain adaptation without supervised contrative learning or applying a staining normalization prior to feature extraction. The adaptation of SCDA to the few-shot learning paradigm shows a considerable increase in classifier performance, being a promising approach in terms of training efficiency. The main limitations are the requirement of labels, the requirement of having the same classes in all hospitals, and the evaluation in a larger multi-hospital scenario ($\text{No. of}\;H>2$). The described limitations open promising avenues of research in terms of adapting the proposed method to cases with no labels and to multi-hospital environments.

\section*{Acknowledgments}
This work has received funding from the Spanish Ministry of Economy and Competitiveness through projects PID2019-105142RB-C21 (AI4SKIN) and PID2022-140189OB-C21 (ASSIST). The work of Rocío del Amor and Pablo Meseguer has been supported by the Spanish Ministry of Universities under an FPU Grant (FPU20/05263) and valgrAI - Valencian Graduate School and Research Network of Artificial Intelligence, respectively. 
\begin{small}
\bibliographystyle{ieeetr}
\bibliography{Comunicacion}
\end{small}

\end{document}